\documentclass[10pt,twocolumn,letterpaper]{article}

\usepackage{cvpr}              
\usepackage{lipsum}
\usepackage{bbding}
\usepackage{tabularx}
\usepackage{multirow}
\usepackage{amsmath,amssymb,stmaryrd}

\usepackage{amssymb}
\usepackage{pifont}

\newcommand{\cmark}{\checkmark} 
\newcommand{\xmark}{\ding{55}}  

\usepackage{xspace}

\newcommand{\dataset}{PARSE-10K\xspace}
\newcommand{\method}{PAG\xspace}

\definecolor{cvprblue}{rgb}{0.21,0.49,0.74}
\usepackage[pagebackref,breaklinks,colorlinks,allcolors=cvprblue]{hyperref}
\usepackage{float}
\usepackage[linesnumbered,ruled,vlined]{algorithm2e}
\usepackage{amssymb}
\usepackage{amsmath}
\usepackage[accsupp]{axessibility}

\def\paperID{} 
\def\confName{CVPR}
\def\confYear{2026}

\title{PARSE: Part-Aware Relational Spatial Modeling}

\author{
    Yinuo Bai\textsuperscript{1,2}\quad
    Peijun Xu\textsuperscript{1}\quad
    Kuixiang Shao\textsuperscript{1}\quad
    Yuyang Jiao\textsuperscript{1}\quad 
    Jingxuan Zhang\textsuperscript{1}\\
    Kaixin Yao\textsuperscript{1,2,$\dag$}\quad
    Jiayuan Gu\textsuperscript{1,*}\quad
    Jingyi Yu\textsuperscript{1,*}\quad
    \\
    \small{\textsuperscript{1}ShanghaiTech University} \quad
    \small{\textsuperscript{2}Deemos Technology}
    \\
    {\tt\small \{baiyn2022, xupj2025, shaokx2025, jiaoyy2022, zhangjx12023,}\\
    {\tt\small yaokx2024, gujy1, yujingyi\}@shanghaitech.edu.cn}
    \\
    \small{\textsuperscript{$\dag$}Project Leader \quad \textsuperscript{*}Corresponding Author}
}

\begin{document}

\twocolumn[{
\maketitle
\begin{figure}[H]
    \hsize=\textwidth
    \centering
    \includegraphics[width=\textwidth]{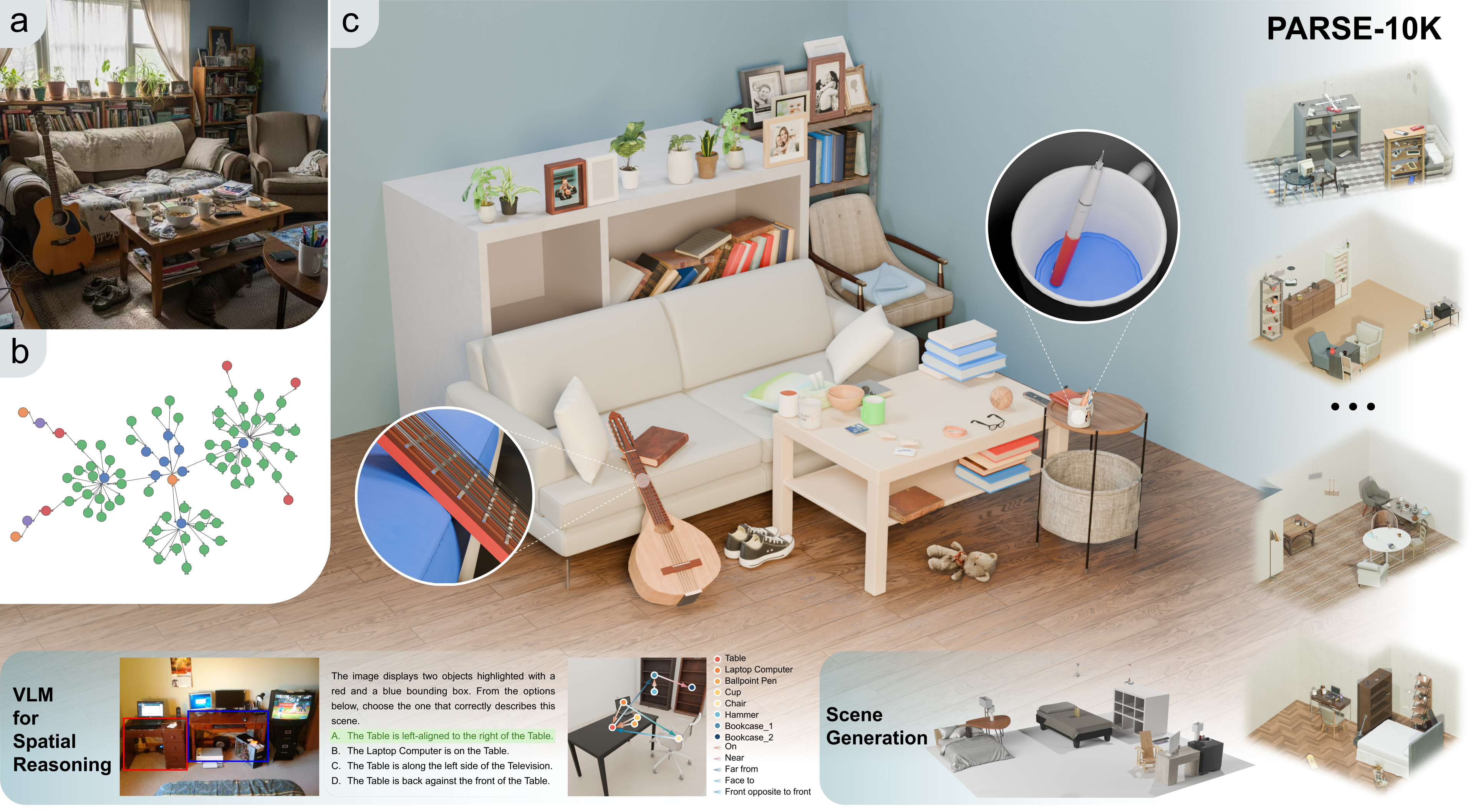}
    \caption{
        \textbf{Overview of the capabilities of PARSE.} Using a real image (a) as reference, we first construct Part-centric Assembly Graphs (PAGs) (b) that capture its spatial organization of objects. Then, by combining PARSE with physics simulation, we generate physically plausible 3D scenes (c) from these PAGs, featuring diverse inter-object relationships and rich part-level contacts. Furthermore, we introduce PARSE-10K, a collection of high-quality 3D indoor scenes with fully part-segmented object instances, which effectively supports downstream tasks such as fine-tuning VLMs for spatial reasoning and enhancing 3D scene generation.
    }
    \label{fig:teaser}
\end{figure}
}]

\begin{abstract}

Inter-object relations underpin spatial intelligence, yet existing representations—linguistic prepositions or object-level scene graphs—are too coarse to specify which regions actually support, contain, or contact one another, leading to ambiguous and physically inconsistent layouts. To address these ambiguities, a part-level formulation is needed; therefore, we introduce PARSE, a framework that explicitly models how object parts interact to determine feasible and spatially grounded scene configurations. PARSE centers on the Part-centric Assembly Graph (PAG), which encodes geometric relations between specific object parts, and a Part-Aware Spatial Configuration Solver that converts these relations into geometric constraints to assemble collision-free, physically valid scenes. Using PARSE, we build PARSE-10K, a dataset of 10,000 3D indoor scenes constructed from real-image layout priors and a curated part-annotated shape database, each with dense contact structures and a part-level contact graph. With this structured, spatially grounded supervision, fine-tuning Qwen3-VL on PARSE-10K yields stronger object-level layout reasoning and more accurate part-level relation understanding; furthermore, leveraging PAGs as structural priors in 3D generation models leads to scenes with substantially improved physical realism and structural complexity. Together, these results show that PARSE significantly advances geometry-grounded spatial reasoning and supports the generation of physically consistent 3D scenes.

\end{abstract}    
\section{Introduction}
\label{sec:intro}

Modeling inter-object relations is the next frontier of spatial intelligence because many fundamental tasks—scene generation~\cite{li2019grains,lin2024instructscene}, layout synthesis~\cite{feng2023layoutgpt}, tidying~\cite{wu2023tidybot}, packing~\cite{zhao2023learning}, stacking~\cite{lee2021beyond}, and embodied manipulation~\cite{jiao2022sequential,zuo2021graph}—depend more on how objects relate than on their isolated shapes. Relations encode support, containment, attachment, occlusion, and accessibility, which determine stability, affordances, and task feasibility. This view resonates with Latour’s actor–network theory (ANT)~\cite{10.1093/oso/9780199256044.001.0001}: objects derive meaning and function from the network of relations they maintain with other objects and agents, not from intrinsic properties alone. 
The critical question we address is \textbf{how to operationalize these rich relations into an effective representation for spatial modeling.}

Vision-language models (VLMs)~\cite{achiam2023gpt,openai2025gpt5,qwen2025qwen3vl} offer a promising path for understanding inter-object relations expressed through prepositions such as \textit{on}, \textit{in}, or \textit{against}. However, \textbf{these linguistic cues are inherently coarse and context-dependent}: ``a book on a table" may refer to the spine or the cover contacting the surface, while ``a guitar leaning on a bookcase" could involve its head or body. 
Such expressions are underspecified regarding contact points or supporting regions, making their translation into spatial configurations fundamentally ambiguous.
This limitation also exists in prior relational representations, most notably scene graphs~\cite{johnson2015image, krishna2017visual}. Prior scene graph representations operate at object-level granularity~\cite{gu2019unpaired, shi2019explainable, johnson2018image, gu2019scene, chang2021comprehensive}, providing insufficient specificity for fine-grained spatial understanding and realistic scene generation.
We argue that \textbf{a more powerful and versatile representation emerges from modeling interactions at the part level}. Part-level relations bridge high-level language descriptions and low-level spatial configurations. For instance, a chair stands on the floor \emph{via its feet}, a mug rests on a table \emph{by its base}, and a broom leans against a wall \emph{at its tip}. This part-centric specification transforms ambiguous prepositions into concrete geometric constraints, effectively pruning the vast search space of valid configurations. When integrated into representations such as scene graphs, these fine-grained relations enable a more structured and controllable approach to spatial reasoning and scene synthesis.

In this work, we propose \textbf{PARSE}, \textbf{PA}rt-aware \textbf{R}elational \textbf{S}patial mod\textbf{E}ling. At its core is the Part-centric Assembly Graph (PAG), a descriptive scene representation where each edge encodes geometric relations between specific parts of connected object nodes. The PAG is organized as a directed acyclic graph, with a hierarchy that guides the assembly of objects into a spatially complex scene. Building on this representation, we introduce a Part-Aware Spatial Configuration Solver, which instantiates PAGs as valid 3D scenes. The solver converts each inter-part relation into geometric constraints, progressively narrowing the feasible pose space of each object and then sampling collision-free solutions efficiently. By traversing the graph from the root, it incrementally generates scenes that adhere to the underlying part-aware structure.

Building on this framework, we construct \dataset, a large and high-quality dataset of 3D indoor scenes with fully part-segmented object instances. We begin by extracting layout priors from real images to obtain a set of semantically plausible and structurally complex PAGs. In parallel, we consolidate multiple public datasets with part annotations~\cite{deitke2023objaverse, collins2022abo, fu20213dfront, wang2025partnext} and incorporate part-segmented generative assets~\cite{zhang2024clay} to build a retrieval database covering 132 object categories for scene assembly. Leveraging these PAGs and the part-level database, we generate 10,000 indoor scenes across 17 room types, each characterized by rich contact structures and accompanied by a corresponding part-level contact graph, which offers an additional source of fine-grained contact information for downstream tasks.

To evaluate the utility of our dataset, we fine-tune Qwen3-VL~\cite{qwen2025qwen3vl} on \dataset and assess its performance on spatial reasoning tasks. The fine-tuned model shows consistent improvements in both object-level layout reasoning and part-level relational understanding. Furthermore, incorporating PAGs from our dataset as structural priors in 3D generation networks significantly enhances the physical realism and structural complexity of synthesized scenes. These results demonstrate the effectiveness of \method in advancing geometry-grounded spatial reasoning and physically consistent 3D scene generation.

\section{Related Work}
\label{sec:relatedwork}

\begin{figure*}[tbp]
    \centering
    \includegraphics[width=1\textwidth]{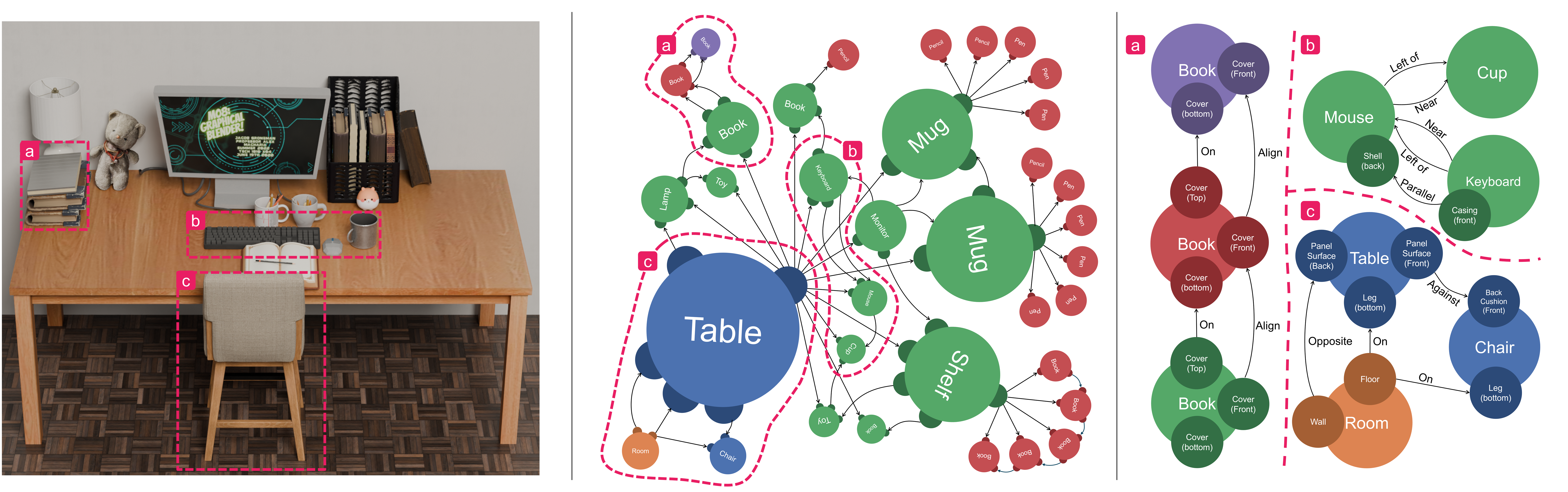}
    \caption{\textbf{An illustrative example of our Part-centric Assembly Graph (PAG).} Left: A 3D scene generated by PARSE, with specific local regions highlighted. Middle: The corresponding global PAG, emphasizing the sub-PAGs that match the highlighted regions on the left. In the graph, large labeled circles represent object nodes, while small dark circles attached to them represent part nodes (unrelated part nodes are omitted for clarity). Right: Zoomed-in views of three sub-PAGs. These panels explicitly annotate the specific surfaces used to define relational constraints, alongside the corresponding Object-Level Spatial Edges and Part-Level Geometric Edges.}
    \label{fig:PAG}
\end{figure*}
Scene graphs~\cite{johnson2015image, krishna2017visual} provide a structured representation of objects and their relations, powering progress in captioning~\cite{yang2019auto, zhong2020comprehensive}, VQA~\cite{antol2015vqa, teney2017graph}, and image retrieval~\cite{schuster2015generating, wang2020cross}. Their extension to 3D~\cite{armeni20193d, kim20193} incorporates geometry and spatial layout, advancing scene understanding~\cite{wald2020learning, wu2021scenegraphfusion} and embodied reasoning~\cite{werby2024hierarchical, li2022embodied}. However, these methods treat objects as indivisible units, leaving them unable to capture the part-level interactions that determine physical stability and support. To overcome this limitation, our PAG models fine-grained part–part relations, enabling explicit reasoning about contact, support, and attachment beyond the capabilities of object-level scene graphs.

Recent works~\cite{cast} in 3D scene generation incorporate inter-object spatial relations to improve structural plausibility. Graph-based approaches condition scene synthesis on semantic or geometric layout structures~\cite{gao2023scenehgn, lin2024instructscene, zhai2023commonscenes, gao2024graphdreamer}, while multimodal and diffusion-based models further align language with 3D geometry to directly produce structured and coherent environments~\cite{tang2024diffuscene, ran2025direct, hollein2023text2room, fu2024anyhome}. Beyond network-based paradigms, procedural generation provides an explicit rule-based mechanism for specifying spatial structure. Early systems such as ProcTHOR~\cite{deitke2022} rely on rigid placement rules that ensure plausible layouts but limit controllability to coarse factors like room type, preventing users from directly specifying inter-object relations. More recent methods improve flexibility by leveraging large language models to map linguistic spatial cues to object placements~\cite{feng2023layoutgpt, sun2025layoutvlm} or to generate constraint programs that can be solved by geometric optimizers~\cite{yang2024holodeck, zhou2025roomcraft}. However, using LLMs as intermediaries introduces semantic ambiguity, often weakening the precision of the resulting geometric constraints. Infinigen~\cite{raistrick2024infinigen} mitigates this by adopting human-readable spatial rules, allowing users to author precise and highly controllable layouts. However, existing procedural systems remain object-centric and cannot capture the fine-grained part interactions needed for precise physical arrangements, leading to inefficient search over large solution spaces. In contrast, our PAG-guided solver encodes explicit part–part constraints, sharply reducing the feasible space and enabling far more efficient generation with higher geometric fidelity and physical consistency.

At the data level, existing indoor scene datasets exhibit analogous limitations. Real-world scanned datasets~\cite{dai2017scannet, chang2017matterport3d, straub2019replica, yeshwanth2023scannet++} provide high-fidelity spatial information and semantic labels, but their object-level meshes are often noisy, incomplete, or fused due to occlusion and reconstruction artifacts, hindering accurate physical reasoning. Synthetic datasets~\cite{fu20213dfront, li2021openrooms, deitke2022, yu2025metascenes} offer cleaner CAD models and greater diversity. However, many meshes are not cleanly decomposed into distinct object instances and lack part-level granularity, making it difficult to reliably model or detect critical, physics-grounded inter-object relations. To fill this gap, PARSE-10K provides consistent part-level annotations and explicit physical relations, delivering the fine-grained supervision absent from existing indoor scene datasets.

\section{Part-Centric Assembly Graph}
\label{sec:pag}

Previous work on scene graphs~\cite{krishna2017visual} often models relationships between whole objects, limiting their precision in capturing fine-grained interactions. To enable a deeper spatial understanding and more precise 3D scene generation, we introduce the Part-centric Assembly Graph (PAG), a representation centered on the expressive power of part-aware relations. As illustrated in \cref{fig:PAG}, the PAG is a hierarchical graph designed to explicitly model the detailed geometric constraints between object parts, providing a structured foundation for both analyzing and synthesizing complex, physically coherent scenes.

 \begin{figure*}[tbp]
    \centering 
    \includegraphics[width=\linewidth]{images/pipe.jpg}
    \caption{\textbf{Controllable Scene Synthesis via Part-Aware Spatial Configuration Solver.}
    \textbf{(a) Coarse Localization:}  The solver first prunes occupied regions (red) from the 2D support surface, then further contracts the feasible space using object-level spatial relations (orange).
    \textbf{(b) Part-Level Alignment:} Precise geometric alignment is achieved by enforcing constraints (\eg, coplanarity) between specific surfaces identified by the solver. This drastically reduces the feasible pose space for final pose sampling.
    \textbf{(c) Fine-Grained Relational Control:} Specifying different part-level geometric relations in the PAG results in distinct and predictable arrangements, showcasing the framework's fine-grained controllability.
    }
    \label{fig:pipeline} 
\end{figure*}

\subsection{Nodes: A Two-Level Structure}
\label{sec:pag_nodes}

To effectively model part-aware relations, the nodes~($\mathcal{V}$) in a PAG are organized into a two-level structure.

\paragraph{Object Nodes ($\mathcal{V}_O$).}
These nodes form the upper level, representing the primary entities in a scene. Each object node encapsulates a semantic query, not a specific 3D instance. This query can be either a single, concrete category or an explicit set of candidate categories. This design defers the choice of a specific geometric instance to the synthesis stage, greatly enhancing the compositional diversity of generated scenes.

\paragraph{Part Nodes ($\mathcal{V}_P$).}
These nodes form the lower level and are the fundamental units for our part-aware approach. Each object node serves as a parent to a set of part nodes representing its geometrically meaningful components (\eg, a ``chair'' connects to its ``legs", ``seat", and ``back cushion").  Each part is further defined by a set of labeled surfaces (\eg, \textit{top, bottom, front, back, left, right}) that are assigned with respect to the asset's canonical pose. These labeled surfaces act as the specific geometric interfaces for alignment and contact, enabling precise constraint definition that govern the scene's assembly structure. 

\subsection{Edges: Part-Aware Relations}
\label{sec:pag_edges}

The edges ($\mathcal{E}$) in a PAG primarily model the rich, part-aware relations for scene assembly. Along with the intra-object edges that associate each object node with its own parts, the PAG mainly encodes inter-object relations through edges at two different granularities:

\paragraph{Object-Level Spatial Edges ($\mathcal{E}_{\text{obj}}$).}
These edges encode coarse spatial relations, such as \textit{left of, behind, near} and \textit{in front of}. Operating at the object level, they connect two object nodes to define macroscopic arrangements. They serve as optional, high-level constraints that guide the overall scene layout.

\paragraph{Part-Level Geometric Edges ($\mathcal{E}_{\text{part}}$).}
These edges encode fine-grained geometric relations, forming the core of the PAG's expressiveness. Specifically, each edge is associated with a spatial preposition (\eg, \textit{on, in, against,} and \textit{aligned with}). Precise physical interactions are specified by connecting two specific part nodes belonging to different parent objects. This part-level linkage enables the graph to encode highly nuanced arrangements. For instance, describing ``a book toppled forward onto a table'' simply requires an \textit{on} edge connecting the book's ``cover'' part node, labeled with the ``front'' surface, to the table's ``surface plane'' part node, labeled with the ``top'' surface.

\begin{table*}[htbp]
    \hsize=\textwidth
    \centering
    \caption{
        \textbf{Comparison of indoor 3D scene datasets.} 
        Columns from left to right denote: dataset name, number of scenes, number of objects, 
        average objects per scene, layout generation method, whether physics simulation or optimization is applied, 
        whether object parts are annotated, and whether part–part contact annotations are provided.
    }
    \resizebox{\textwidth}{!}{
        \begin{tabular}{lccccccc}
            \toprule
            Dataset & \# Scenes & \# Objects & \# Avg.Objects &  Layout Generation & Physical Optimization & Part Annots. & Part-Level Contact Annots.  \\
            \midrule
            HSSD-200~\cite{khanna2024habitat}            & 211              & 18656   & 329.7                   & Human-designed                             & \xmark    & \xmark & \xmark \\
            3D-FRONT~\cite{fu20213dfront}          & 18968              & 13151   & 6.9                   & Human-designed                            & \xmark    & \xmark & \xmark \\
            FurniScene~\cite{zhang2024furniscene}    & 111698              & 39691   & 14.4                   & Human-designed                            & \xmark    & \xmark & \xmark \\
            METASCENES~\cite{yu2025metascenes}            &706              & 15366   & -                   & Real-world Scanned                             & \cmark    & \xmark  & \xmark\\
            \midrule
            PARSE-10K (Ours) & 10000              & 17372   & 49.9                   & Real-image Guided                             & \cmark    & \cmark & \cmark \\
        \bottomrule
        \end{tabular}
    }
    \label{tab:dataset_comparison}
\end{table*}

\subsection{Hierarchical Assembly Structure}
\label{sec:pag_structure}

A static 3D scene can be viewed as a collection of dense, interdependent geometric relations. To manage this complexity, our PAG representation adopts an assembly-centric perspective, viewing a stable scene as the outcome of a sequential construction process. This view allows us to represent scene structure in a more computationally tractable manner.

This assembly-centric perspective is realized through a key structural constraint of PAG: the entire graph must be a Directed Acyclic Graph (DAG). This global acyclic property is the necessary mathematical structure for representing a sequential process without circular dependencies, and it directly ensures physical realizability by enforcing a valid, step-by-step construction order. Additionally, we define that each object must have a unique physical supporter, a rule that naturally organizes the scene into a clear hierarchical structure. Ultimately, this overall design makes the scene-wide constraint satisfaction problem computationally tractable by decomposing it into a well-defined sequence of localized subproblems—one for each object in the assembly order.

\section{PARSE-10K}
\label{sec:contactnet}

We introduce the PARSE framework, our procedural synthesis pipeline that instantiates abstract PAGs into physically plausible and geometrically precise 3D scenes. This framework serves as the engine for building \dataset, a large-scale dataset of diverse, part-aware indoor scenes. At the core of our framework is the Part-Aware Spatial Configuration Solver.

\subsection{Part-Aware Spatial Configuration Solver}\label{constraint-solver}

Given a PAG, the Part-Aware Spatial Configuration Solver instantiates it into a 3D scene by processing its object nodes in a topological sort. This traversal follows the sequential assembly order induced by the PAG's support relations. For each object in this sequence, the solver finds a valid pose through a coarse-to-fine process of progressive refinement. As illustrated by the key steps in \cref{fig:pipeline}, it sequentially applies all relevant constraints, with each new constraint further shrinking the object's feasible pose space until a precise solution is found. The instantiation of each object node unfolds as follows:

\paragraph{Coarse Localization.}
As each object node in a PAG has a unique supporter, the solving process begins within a 2D candidate region defined on the support surface, from which all previously occupied areas have been excluded. We first apply the high-level, object-level spatial edges. These constraints contract the node's feasible region to a smaller subspace. For instance, a ``\textit{left of}" relation imposes a plane that restricts the object's valid translational range to one side of the target object.

\paragraph{Part-Level Alignment.}

At this stage, a specific 3D asset, complete with per-part segmentation and semantic labels, is instantiated from our asset library based on the node's semantic query. Once an asset is chosen, the solver resolves the part-level geometric constraints. Guided by the spatial preposition of the connecting edge, the resolution strategy diverges based on surface specifications. If specific labeled surfaces of the connected part nodes are explicitly provided, the solver directly uses these identifiers. If exact surfaces are not specified, the solver performs a geometric reasoning step. For example, for an \textit{on} relation, it dynamically identifies the supported part's lowest bottom surface while searching the target part for a suitable upward-facing support plane. The identified surfaces—whether explicitly provided or geometrically inferred—are then used to formulate a new set of geometric constraints. These constraints typically enforce properties such as making the two surfaces parallel and bringing them into contact. Each new constraint, solved in conjunction with existing ones, further contracts the object's feasible pose space towards a minimal valid subspace.

\begin{figure*}[ht]
    \hsize=\textwidth
    \centering
    \includegraphics[width=1\textwidth]{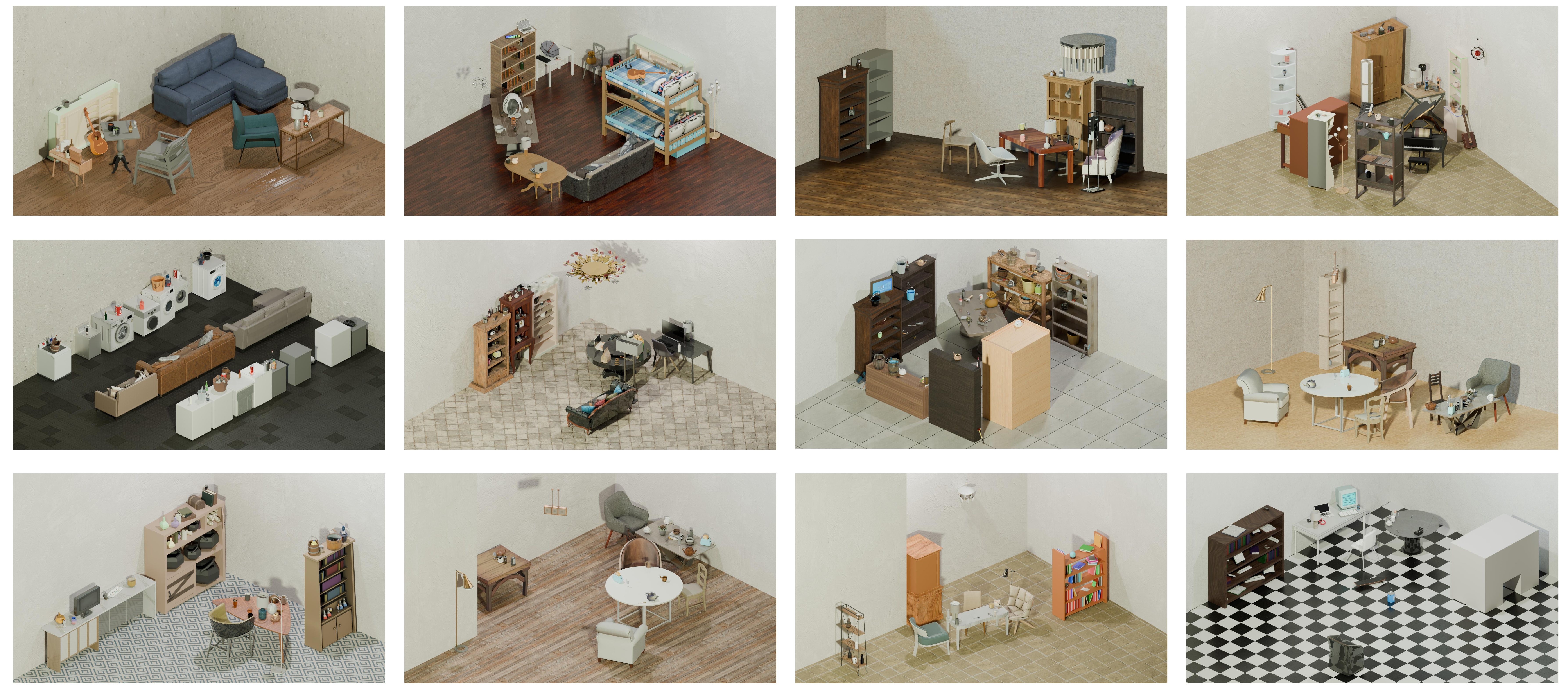}
    \caption{
        \textbf{Gallery of PARSE-10K.} 
    }
    \label{fig:gallery}
\end{figure*}

\paragraph{Final Pose Sampling and Validation.}
Once all constraints have been applied, we randomly sample a final pose from this subspace and validate it for 3D collisions and physical-semantic plausibility (\eg, for an \textit{in} relation, we validate the degree of enclosure via multi-directional ray-casting). Because our solving process is a deterministic accumulation of constraints, any pose sampled from this final subspace is guaranteed a priori to satisfy all non-collision-related geometric and spatial relations. This ensures a high success rate for the final validation step, avoiding costly cycles of blind rejection sampling.

To ensure physical plausibility, the fully instantiated scene undergoes a final refinement step via a brief dynamic simulation in Sapien~\cite{xiang2020sapien}. This process yields a final 3D scene with an enhanced level of physical realism and stability. From this stable configuration, we additionally generate a detailed part-level contact graph by identifying all part pairs in close proximity (\eg, $\leq$1mm).

\subsection{Dataset Statistics and Analysis}\label{dataset-statistics}
Leveraging the PARSE framework's explicit modeling of part-level spatial relations, we construct PARSE-10K, a large-scale dataset comprising \textbf{10,000} unique indoor scenes, each annotated with a corresponding part-level contact graph. The dataset's compositional diversity is rooted in its rich asset library, which contains over \textbf{17,372} part-segmented and semantically-labeled assets across \textbf{132} object categories. Each scene is densely populated with an average of \textbf{49.9} objects, resulting in a high degree of physical plausibility and rich, part-level relational complexity. As showcased in \cref{fig:gallery}, this explicit modeling enables the generation of intricate arrangements—such as precisely stacked objects, items leaning against surfaces, and complex container-content relationships—that are difficult to synthesize or annotate in existing datasets.
Our comparative analysis, detailed in \cref{tab:dataset_comparison}, positions PARSE-10K uniquely within the landscape of 3D scene datasets. 
PARSE-10K bridges this fundamental gap, providing a large-scale resource of scenes that are simultaneously physically grounded, compositionally diverse, and richly annotated with part-aware geometric relations.

\section{Experiments on Spatial Tasks}
\label{sec:experiments}

This section demonstrates the broad utility of PARSE-10K in spatial understanding and generative tasks. First, leveraging its rich spatial relations and fine-grained part-level contacts, we benchmark state-of-the-art VLMs and propose targeted improvements to their spatial grounding and contact reasoning (\cref{Spatialvlm}). Second, the dataset’s densely annotated, relation-rich scenes serve as a rigorous testbed for controllable and fidelity-preserving scene synthesis (\cref{genscene}).

\subsection{VLM for Spatial Reasoning}
\label{Spatialvlm}

\begin{table*}[ht]
\centering
\caption{\textbf{Quantitative comparison with baselines.} We evaluate models across three tasks: visual relation MCQ, part-level contact MCQ, and scene graph generation (SGG). 
For SGG, each metric is reported in the format \textit{With BBox Matching / No BBox Matching}. The \textit{Avg.} column indicates the average number of relations generated per scene by each model.}
\small
\resizebox{0.9\linewidth}{!}{
\begin{tabular}{lcccccc}
\toprule
\multirow{2}{*}{Models} & \multirow{2}{*}{Visual Relation$\uparrow$} & \multirow{2}{*}{Part-level Contact$\uparrow$} & \multicolumn{4}{c}{Scene Graph Generation} \\
\cmidrule(lr){4-7}
 &  &  & Recall$\uparrow$ & Precision$\uparrow$ & F1 Score$\uparrow$ & Avg. \\
\midrule
GPT-5           & 82.1 & 75.2 & 13.7/40.9 & 13.9/41.3 & 13.8/41.1 & 15.3 \\
Gemini-2.5-Pro  & 85.0 & 75.6 & 40.5/43.4 & 48.6/52.0 & 44.2/47.3 & 12.9\\
Claude-Opus-4   & 80.3 & 73.2 & 8.0/33.7  & 12.7/53.7 & 9.8/41.4  & 9.7 \\
Robobrain2.0    & 60.8 & 37.2 & 9.2/11.3  & 26.7/32.8 & 13.7/16.9 & 5.6 \\
Qwen3-VL        & 86.2 & 60.4 & 26.0/29.6 & 46.0/52.4 & 33.2/37.9 & 8.7 \\
Ours            & \textbf{97.4} & \textbf{86.2} & \textbf{73.2/74.8} & \textbf{80.3/82.0} & \textbf{76.6/78.2} & 14.1 \\
\bottomrule
\label{tab:comparison}
\end{tabular}
}
\end{table*}

\textbf{Dataset construction.}
We synthesize a large collection of rendered scene images paired with part-level and object-level relation graphs. For each scene, we render multiple camera views and extract the subset of the scene graph corresponding to the objects and parts visible in that view. From these annotations, we construct three evaluation tasks. (1) \textit{Visual Relation Multiple Choice Questions (MCQ)}: for a sampled relation triplet (two objects and their relation), we mark the objects on the image and present a multiple choice question, following protocols from several spatial-understanding benchmarks~\cite{Du2024EmbSpatialBenchBS, zhou2025roborefer}. Distractors are created by randomly replacing object and relation labels using a predefined object and relation vocabulary. (2) \textit{Part-level Contact MCQ}: for a sampled visible part–part contact pair, we form a multiple choice question of the form “Part M of Object A contacts Part N of Object B”; the two objects are marked on the image, and distractors are generated by randomly replacing objects and parts using a prebuilt object-part mapping. (3) \textit{Scene Graph Generation (SGG)}: given an image and the full set of candidate object names and relation types, the model must both localize all objects (2D bounding boxes and labels) and enumerate the relations among them; object entries contain the label and 2D bbox, and relations are reported as triplets.

\begin{figure}
    \centering
    \includegraphics[width=1\linewidth]{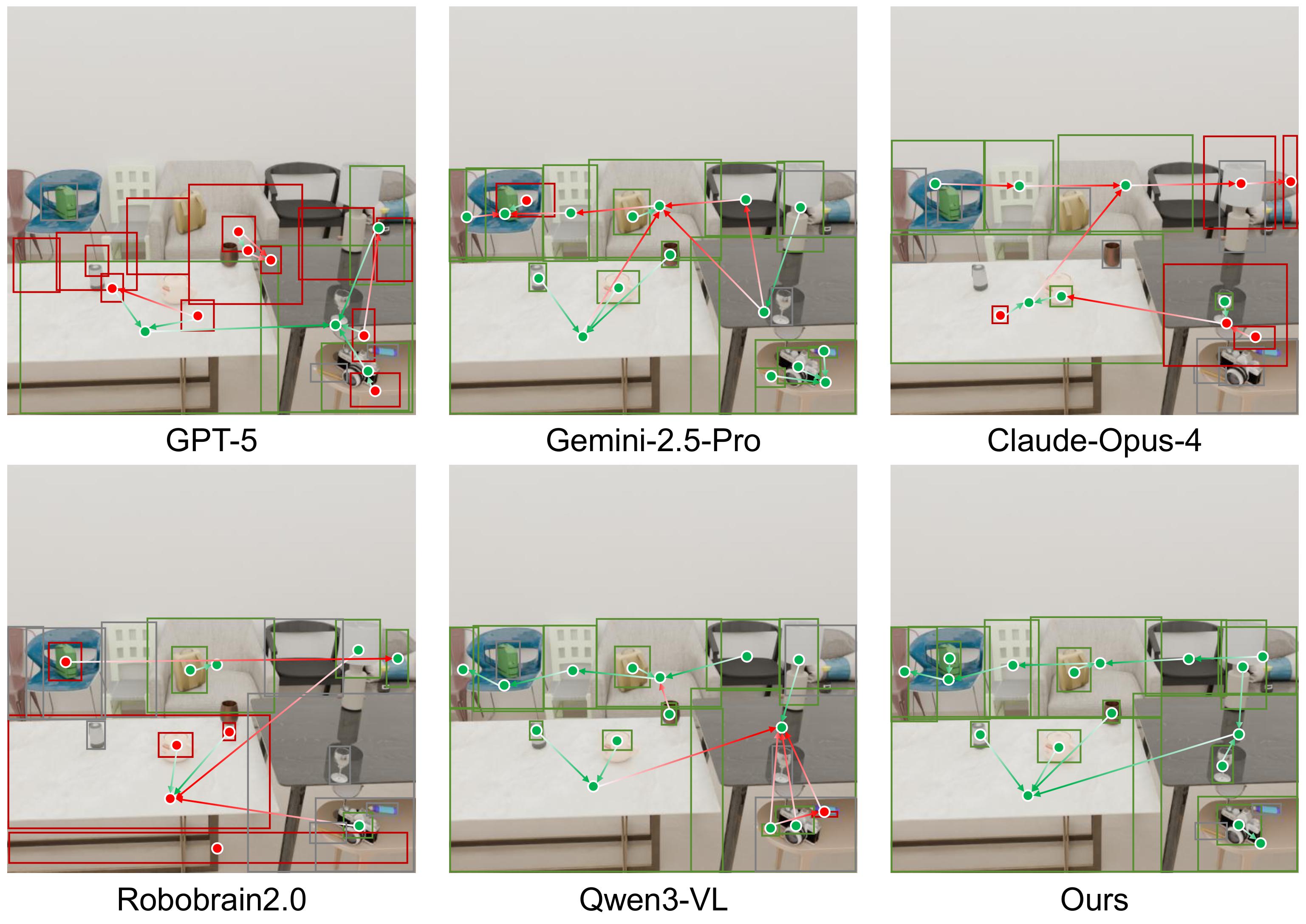}
    \caption{\textbf{Visualization of model-predicted graphs.} Green boxes indicate objects correctly matched by both label and grounding; red boxes indicate failed matches; gray boxes denote missed detections. Green arrows denote relations judged correct under the grounding-agnostic metric, red arrows denote incorrect relations.}
    \label{fig:sg}
\end{figure}

\textbf{Experimental setup.}
We evaluate several leading VLMs as baselines: GPT-5~\cite{openai2025gpt5}, Gemini-2.5-Pro~\cite{Comanici2025Gemini2P}, Claude-Opus-4~\cite{anthropic2025claudeopus4}, Robobrain2~\cite{RoboBrain2.0TechnicalReport}, and Qwen3-VL~\cite{qwen2025qwen3vl}. These are compared against our model (denoted ``Ours''), which is fine-tuned from Qwen3-VL on our constructed dataset to study the efficacy of the targeted data. For the two MCQ tasks, we use accuracy on the selected option as the metric. To verify the model's generalization ability, we also add some manually labeled real images from the COCO~\cite{Lin2014MicrosoftCC} dataset to the test set. For Scene Graph Generation, we first normalize synonyms in model outputs, then perform class-wise matching between predicted and ground-truth boxes using an IoU-based Hungarian assignment~\cite{Kuhn1955TheHM}. 
A predicted relation is considered correct only if it connects the correctly matched object instances. Since models vary in the volume of relations they generate, we report the average number of predicted relations alongside precision, recall, and F1 score to provide a comprehensive evaluation.

\textbf{Results and analysis.}
\cref{tab:comparison} summarizes the quantitative results on the MCQ tasks and aggregated part-contact scores. On the Visual Relation MCQ task, our fine-tuned model achieves 97.4\%, substantially outperforming the baselines. On the Part-level Contact MCQ task, our model likewise leads with 86.2\%. In the Scene Graph Generation task, the fine-tuned model substantially outperforms all baselines: it yields marked gains in object recognition, 2D localization, and the annotation of spatial relations. By explicitly training on PARSE-10K’s dense, part-level supervision the model produces more complete and more accurately grounded relation sets compared to generalist VLMs. In contrast, models such as GPT-5 and Claude—while strong at high-level relational reasoning—exhibit weaker visual grounding and therefore suffer during the bbox-matching stage, which degrades their downstream relation scores. As an additional analysis, we also report relation accuracy under a grounding-agnostic metric (i.e., measuring relation correctness without requiring bbox matches) to separate pure relational reasoning from grounding performance. \cref{fig:sg} visualizes the model-predicted graphs.

Our experiments demonstrate that fine-tuning on the constructed PARSE-10K dataset significantly enhances both visual grounding and relational reasoning. The fine-tuned model achieves the highest performance across all tasks, with notable improvements in MCQ accuracy and a substantial lead in Scene Graph Generation metrics. Compared to general-purpose VLMs such as GPT-5, Gemini-2.5-Pro, and Claude-Opus-4, our model produces more complete and accurately grounded scene graphs. The additional grounding-agnostic evaluation further confirms that the observed gains stem not only from improved visual localization but also from stronger relational understanding.

\subsection{Scene Generation}
\label{genscene}

\textbf{Dataset construction.}
The goal of the scene generation task is to generate rotation, translation, and scale for each given object, with or without scene graph control, and then combine them into a reasonable scene. PARSE-10K poses particular challenges for this task: scenes contain many objects, exhibit complex hierarchical relationships, and require precise object–object contacts. To capture geometric information, we encode each mesh with a Michelangelo~\cite{zhao2023michelangelo} encoder and feed the resulting per-object geometry features to the network. We encode the PAG using CLIP~\cite{Radford2021LearningTV} and convert its output into a relation embedding matrix. For training targets, we extract object poses from simulated scenes and use them as the denoising targets for the diffusion model.

\textbf{Experimental setup.}
We build a graph-transformer-based diffusion network inspired by InstructScene~\cite{lin2024instructscene}. At each denoising layer, the model fuses the object geometry features with the current noisy pose via cross-attention; the scene-graph control is injected into attention layers using a FiLM-style~\cite{perez2018film} modulation so that relational constraints can influence the pose refinement. We train and evaluate both conditioned and unconditioned variants (i.e., with and without PAG control) and follow standard diffusion schedules; detailed training hyperparameters and optimization schedules are provided in the Appendix. We present a qualitative comparison between scenes generated by the state-of-the-art method, InstructScene, trained on the 3D-FRONT~\cite{fu20213dfront} dataset, and those produced by our proposed method trained on the PARSE-10K dataset, under both PAG-conditioned and unconditioned settings. Furthermore, we conduct a user study to quantitatively evaluate the generated scenes in terms of their complexity, realism, and contact plausibility. A total of 20 participants were involved in the study, each evaluating 12 rendered scenes by selecting the one that best fit the given criterion.

\begin{figure}[tbp]
    \centering 
    \includegraphics[width=\linewidth]{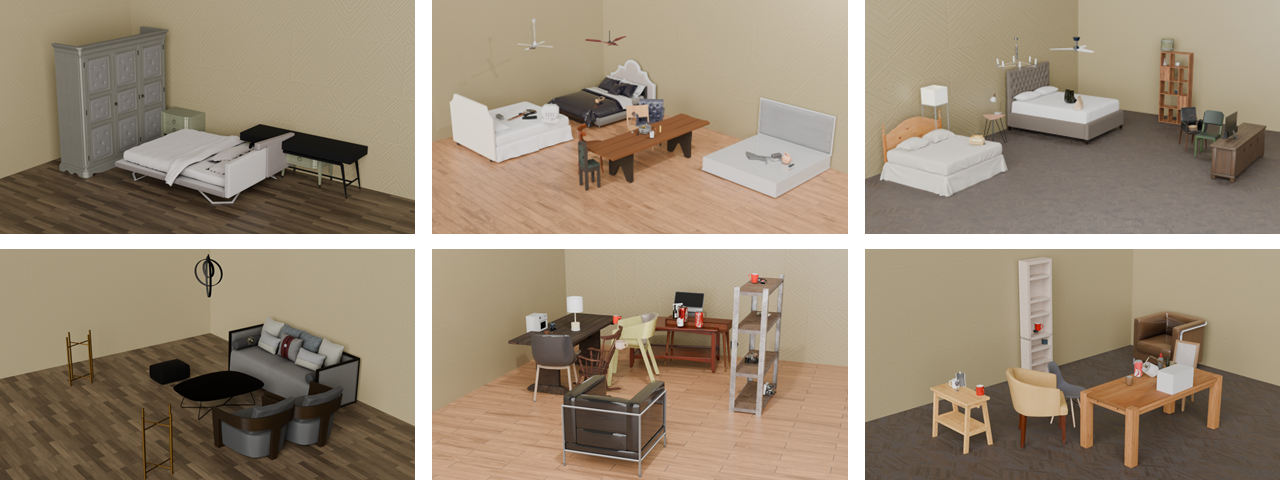}
    \caption{\textbf{Scene generation comparison}. Left column: scenes generated by InstructScene trained on 3D-FRONT. Middle column: scenes generated by our method trained on PARSE-10K without PAG control. Right column: scenes generated by the model with PAG control.}
    \label{fig:genscene} 
\end{figure}

\textbf{Results and analysis.}
Qualitative comparisons (\cref{fig:genscene}) show that training on PARSE-10K produces scenes with a higher object count and richer, more complex contacts than the baseline trained on 3D-FRONT. Conditioning on the scene graph yields scenes whose inter-object relations are more semantically coherent and physically plausible. A user study, summarized in \cref{tab:userstudy}, further quantifies these improvements. Owing to the high complexity and rich contact relationships of our dataset, learning its distribution without PAG conditioning often leads to unrealistic physics and unreasonable layout. Consequently, participants showed limited preference for scenes generated without PAG conditioning. Nevertheless, when conditioned on PAG, the model is able to generate scenes with a larger number of objects and finer contact details. Participants consistently preferred our PAG-conditioned PARSE-10K-trained models on measures of scene complexity, realism, and contact fidelity. 

In summary, our experiments show that the proposed PARSE-10K dataset facilitates the generation of more complex and realistic scenes. Both qualitative comparisons and quantitative user studies confirm that models trained on PARSE-10K produce scenes with higher object counts, richer contact relationships, and greater semantic and physical plausibility than those trained on previous datasets. These results demonstrate the value of our dataset in advancing contact-rich 3D scene-generating techniques.

\begin{table}[tbp]
    \centering
    \caption{\textbf{User study.} 
    The table reflects the percentage of user votes for scenes generated from the corresponding model.}
    \begin{tabularx}{\linewidth}{
        p{1.8cm}
        >{\centering\arraybackslash}X
        >{\centering\arraybackslash}X
        >{\centering\arraybackslash}X
    }
        \toprule
        Method & Complexity$\uparrow$ & Realism$\uparrow$ & Contact Fidelity$\uparrow$\\
        \midrule
        InstructScene & 7.5\% & 33.8\% & 28.8\% \\
        Ours(uncond) & 45.0\% & 27.5\% & 26.3\% \\
        Ours(cond) & \textbf{47.5\%} & \textbf{38.8\%} & \textbf{45.0\%} \\
        \bottomrule
    \end{tabularx}
    \label{tab:userstudy} 
\end{table}

\section{Conclusion}
\label{sec:conclusion}

We introduce PARSE, a part-centric framework that encodes geometric interactions between object parts through a Part-centric Assembly Graph and a Part-Aware Spatial Configuration Solver, enabling the synthesis of physically consistent 3D layouts. We also construct PARSE-10K, a large-scale dataset with dense part-level contact annotations that enhance spatial reasoning and 3D scene generation.

While PARSE and PARSE-10K advance part-level spatial modeling, several limitations remain. Relation definitions are complex and require part-specific coordinate reasoning, making PAG construction partially manual and sensitive to canonical poses. Future work will focus on learning part–part relations directly from geometry, developing more flexible contact representations, expanding the diversity of PARSE-10K, and integrating PARSE into embodied tasks for part-level planning and physically grounded manipulation.
\section*{Acknowledgments}
This work was supported in part by the National Natural Science Foundation of China under Grant W2431046, National Key R\&D Program of China 2025YFA1309603, Central Guided Local Science and Technology Foundation of China YDZX20253100001001,and by MoE Key Lab of Intelligent Perceptionand Human-Machine Collaboration (ShanghaiTech University), the Shanghai Frontiers Science Center of Human-centered Artificial Intelligence. The experiments of this work were supported by HPC Platform of ShanghaiTech University.

{
    \small
    \bibliographystyle{ieeenat_fullname}
    \bibliography{main}
}

\clearpage
\setcounter{page}{1}

\maketitlesupplementary
\setcounter{section}{0}
\renewcommand{\thesection}{\Alph{section}}
\renewcommand{\thesubsection}{\thesection.\arabic{subsection}}

\section{Details of PARSE Framework}
\begin{figure}[h!]
    \centering
    \includegraphics[width=\linewidth]{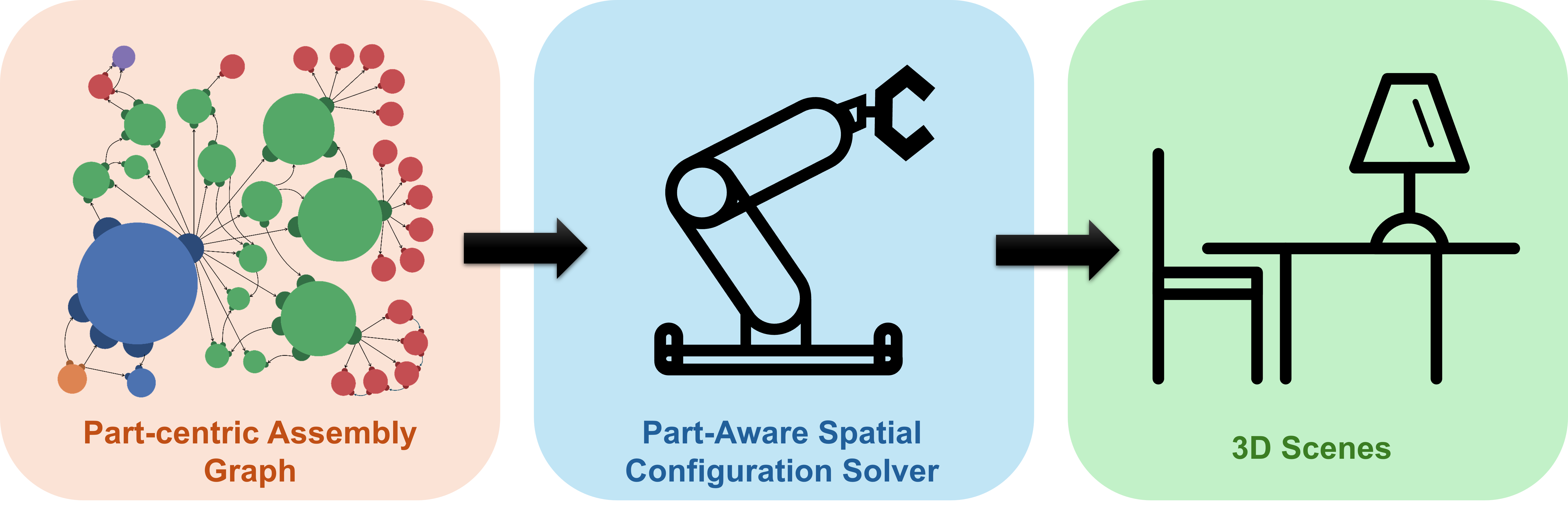}
    \caption{Overview of PARSE Framework}
    \label{fig:ov}
\end{figure}

The overview of the PARSE framework is shown in \cref{fig:ov}, and the specific details of this framework will be elaborated in this section.

\subsection{PAG Representation}
\label{subsec:pag_rep}

\paragraph{Graph Structure and Data Definition.}
We implement the Part-centric Assembly Graph (PAG) as a hierarchical directed acyclic graph $\mathcal{G} = (\mathcal{V}, \mathcal{E})$.

    Nodes ($\mathcal{V} = \mathcal{V}_O \cup \mathcal{V}_P$): Each node $u \in \mathcal{V}_O$ represents a distinct object instance in the scene. In our data structure, $u$ stores global attributes: the semantic category label $C$, a scaling factor $S$, and a retrieved asset $\mathcal{M}$. Each node $v \in \mathcal{V}_P$ serves as a part node attached to a parent object node. 
    Implementation-wise, the part node stores a description tuple $\tau = (l, s)$. Here, $l$ is a semantic label (\eg, “seat”, “leg”, or a default “root”), and $s$ denotes the six directional faces of the part mesh (top, bottom, left, right, front, back), which serve as the primary entities processed by the solver when solving part-level constraints (details discussed later).
    
    Edges ($\mathcal{E} = \mathcal{E}_{obj} \cup \mathcal{E}_{part}$): Object-Level Edges ($\mathcal{E}_{obj}$) are edges that connect two object nodes to enforce macroscopic layout constraints (\eg, \textit{left of}, \textit{near}), primarily used for coarse region proposal. Part-Level Edges ($\mathcal{E}_{part}$) are edges that are stored as a triplet $e = (v_{src}, v_{tgt}, r)$, connecting a source part node $v_{src}$ to a target part node $v_{tgt}$ via a specific spatial relation type $r$.

\paragraph{Spatial Relations.}
The geometric nature of the connection is determined by the relation type $r$. To handle the complexity of 3D spatial interactions, we define a comprehensive dictionary of relations. \cref{tab:relation_definitions} enumerates these relations, categorizing them by their physical support roles and spatial granularity, and detailing the underlying geometric constraints enforced by the solver.

\begin{table*}[ht]
\centering
\small
\renewcommand{\arraystretch}{1.4}
\caption{\textbf{Dictionary of Spatial Relations in PARSE.} We categorize relations into two primary granularities: \textbf{Part-Level} (fine-grained geometric control) and \textbf{Object-Level} (coarse layout). Specific Part-Level relations function as Support Relations, defining the scene's spanning tree. To handle unspecified inputs, we define \textbf{Default Surfaces} for implicit resolving. Notation: $A$ is Child, $B$ is Parent; $S$ denotes surface, $\mathbf{n}$ denotes normal, $\mathbf{x}$ denotes centroid.}
\label{tab:relation_definitions}
\resizebox{\textwidth}{!}{%
\begin{tabular}{@{}l l c l p{7.5cm}@{}}
\toprule
\textbf{Relation ($r$)} & \textbf{Semantic Description} & \textbf{Support?} & \textbf{Default Surfaces ($S_A \to S_B$)} & \textbf{Geometric Constraints \& Solver Logic} \\
\midrule

\multicolumn{5}{l}{\textit{\textbf{1. Part-Level: Support Relations}}} \\
\multicolumn{5}{l}{\textit{These relations define the primary parent-child dependency, organizing the scene assembly order.}} \\

\textbf{\texttt{ON}} & Object rests stably on parent. & Yes & 
Bottom $\to$ Top & 
Contact: $\text{dist}(S_A, S_B) \approx 0$. \newline
Orientation: $\mathbf{n}_A \cdot \mathbf{n}_B \approx -1$ (Anti-parallel). \newline
Stability: $Proj_{xy}(\text{CoM}_A) \in \text{Polygon}(S_B)$. \\

\textbf{\texttt{IN}} & Object is contained inside parent. & Yes & 
Bottom $\to$ Top(internal support surface) & 
Contact: $\text{dist}(S_{A}, S_{B}) \approx 0$. \newline
Inclusion: $\text{RayCastEnclosure}(A, B) > \tau_{contain}$. \\

\textbf{\texttt{HANGING\_BELOW}} & Object suspended under parent. & Yes & 
Top $\to$ Bottom & 
Contact: $\text{dist}(S_A, S_B) \approx 0$. \newline
Orientation: $\mathbf{n}_A \cdot \mathbf{n}_B \approx -1$. \\

\textbf{\texttt{SECURED\_TO}} & Rigidly attached to side/wall. & Yes & 
Back $\to$ Front & 
Contact: $\text{dist}(S_A, S_B) \approx 0$. \newline
Orientation: $\mathbf{n}_A \cdot \mathbf{n}_B \approx -1$. \\

\midrule

\multicolumn{5}{l}{\textit{\textbf{2. Part-Level: Fine-Grained Geometric Constraints}}} \\
\multicolumn{5}{l}{\textit{Fine-grained surface constraints refining the pose without defining assembly hierarchy.}} \\

\textbf{\texttt{AGAINST}} & Stable surface contact. & No & 
Front $\to$ Front & 
Contact: $\text{dist}(S_A, S_B) \approx 0$. \newline
Orientation: $\mathbf{n}_A \cdot \mathbf{n}_B \approx -1$. \\

\textbf{\texttt{LEANING\_AGAINST}} & Leans on parent (contact + angle). & No & 
Back $\to$ Front & 
Contact: $\text{dist}(S_A, S_B) \approx 0$. \newline
Angle: $\mathbf{v}_{up} \cdot \mathbf{v}_{gravity} \approx \cos(\theta), \theta \in (0, \frac{\pi}{2})$. \\

\textbf{\texttt{ALIGNED\_WITH}} & Surfaces flush, facing same dir. & No & 
Front $\to$ Front & 
Coplanarity: $\text{dist}(\text{Plane}(S_A), \text{Plane}(S_B)) \approx 0$. \newline
Orientation: $\mathbf{n}_A \cdot \mathbf{n}_B \approx 1$ (Parallel). \\

\textbf{\texttt{OPPOSITE\_TO}} & Surfaces flush, back-to-back. & No & 
Front $\to$ Front & 
Coplanarity: $\text{dist}(\text{Plane}(S_A), \text{Plane}(S_B)) \approx 0$. \newline
Orientation: $\mathbf{n}_A \cdot \mathbf{n}_B \approx -1$ (Anti-parallel). \\

\textbf{\texttt{PARALLEL\_TO}} & Planes parallel (any distance). & No & 
Front $\to$ Front & 
Orientation: $\mathbf{n}_A \cdot \mathbf{n}_B \approx 1$ (Parallel).  \\

\midrule

\multicolumn{5}{l}{\textit{\textbf{3. Object-Level: Coarse Layout}}} \\
\multicolumn{5}{l}{\textit{Region-based constraints acting on object centroids or bounding boxes.}} \\

\textbf{\texttt{SIDE\_OF}} & Generic half-space constraint. & No & 
N/A & 
Region: $(\mathbf{x}_A - \mathbf{x}_B) \cdot \mathbf{n}_{side} > 0$. \\


\textbf{\texttt{ALONG}} & Arranged along a linear strip. & No & 
N/A & 
Region: $0 < (\mathbf{x}_A - \mathbf{x}_B) \cdot \mathbf{n}_{side}^{\perp} < \delta_{width}$. \\

\textbf{\texttt{NEAR / FAR\_FROM}} & Proximity constraint. & No & 
N/A & 
Distance: $\|\mathbf{x}_A - \mathbf{x}_B\|_2 \lessgtr \tau_{thresh}$. \\

\textbf{\texttt{TOWARD}} & Directional orientation. & No & 
N/A & 
LookAt: $\mathbf{n}_{A, front} \cdot \frac{\mathbf{x}_B - \mathbf{x}_A}{\|\mathbf{x}_B - \mathbf{x}_A\|} \approx 1$. \\

\bottomrule
\end{tabular}%
}
\end{table*}

\subsection{Part-Aware Spatial Configuration Solver Details}
\label{subsec:solver_details}
The detailed algorithmic workflow of our solver, corresponding to the process described below, is formally presented in \cref{alg:solver}.

\begin{algorithm}[t]
\small
\SetAlgoLined
\LinesNumbered
\DontPrintSemicolon

\SetKwInOut{Input}{Input}
\SetKwInOut{Output}{Output}
\SetKwFunction{TopoSort}{TopologicalSort}
\SetKwFunction{SelectAsset}{SelectAsset}
\SetKwFunction{Resolve}{ResolveSurface}
\SetKwFunction{Clip}{SurfaceClip}
\SetKwFunction{SolveRot}{SolveRotation}
\SetKwFunction{Sample}{Sample}
\SetKwFunction{GetPartEdges}{GetPartEdges}
\SetKwFunction{GetObjEdges}{GetObjectEdges}
\SetKwFunction{Check}{ValidatePose}
\SetKwFunction{Occupied}{OccupiedFootprints}
\SetKwFunction{ExtractNormals}{ExtractNormalConstraints}
\SetKwFunction{GetSupp}{GetSupportSurface}
\SetKwFunction{GetPartRegion}{GetPartConstraintRegion}
\SetKwFunction{GetObjRegion}{GetObjectConstraintRegion}

\caption{Coarse-to-Fine Spatial Solver}
\label{alg:solver}

\Input{Graph $\mathcal{G}$, Asset Database $\mathbb{D}$}
\Output{Selected Assets $\mathcal{M}$, Scene Poses $\mathcal{T} = \{(R_i, t_i)\}$}

$\mathcal{L} \leftarrow$ \TopoSort{$\mathcal{G}$}\;
\ForEach{Object Node $v_i \in \mathcal{L}$}{
    
    $\mathcal{M}_i \leftarrow$ \SelectAsset{$v_i, \mathbb{D}$}\;

    $\mathcal{S}_{supp} \leftarrow v_{i}.$\GetSupp{}\;
    $Obs \leftarrow$ \Occupied{$\mathcal{T}$}\; 
    $\mathcal{S}_{free} \leftarrow$ \Clip{$\mathcal{S}_{supp}, Obs$}\;
    
    \ForEach{$e_{obj} \in$ \GetObjEdges{$v_i$}}{
        $Region \leftarrow$ \GetObjRegion{$e_{obj}$}\;
        $\mathcal{S}_{free} \leftarrow$ \Clip{$\mathcal{S}_{free}, Region$}\;
    }
    
    $\Omega_{parts} \leftarrow \emptyset$\;
    \ForEach{$e_{part} \in$ \GetPartEdges{$v_i$}}{
        $S_{child} \leftarrow$ \Resolve{$e_{part}.child$}\;
        $S_{parent} \leftarrow$ \Resolve{$e_{part}.parent$}\;
        $\Omega_{parts}.\text{add}((S_{child}, S_{parent}, e_{part}.relation))$\;
    }
    $\mathcal{C}_{rot} \leftarrow$ \ExtractNormals{$\Omega_{parts}$}\;
    $R \leftarrow$ \SolveRot{$\mathcal{C}_{rot}$}\;
    
    \ForEach{$(S_{child}, S_{parent}, r) \in \Omega_{parts}$}{
        $S_{child}' \leftarrow R \cdot S_{child}$\;
        $Region \leftarrow$ \GetPartRegion{$S_{parent}, S_{child}', r$}\;
        $\mathcal{S}_{free} \leftarrow$ \Clip{$\mathcal{S}_{free}, Region$}\;
    }
    
    \For{$k \leftarrow 1$ \KwTo $MaxTrials$}{
        $t \leftarrow$ \Sample{$\mathcal{S}_{free}$}\;
        \If{\Check{$R, t, \mathcal{T}, \Omega_{parts}$}}{
            $\mathcal{T}[v_i] \leftarrow (R, t)$\;
            \textbf{break}\;
        }
    }
}
\Return $\mathcal{T}$
\end{algorithm}

\paragraph{Asset Instantiation and Geometric Resolution.}
Before any spatial reasoning occurs, the abstract PAG nodes must be grounded into 3D assets.
\begin{itemize}
    \item \textbf{Asset Retrieval and Scaling:} For the current object node $u_i$, we retrieve a specific 3D asset $\mathcal{M}_i$ from the database $\mathbb{D}$ that matches the node's semantic category. To ensure physical realism, we define a scale range for each object category and, for every instance, sample its scale from the corresponding category-specific range.
    
    \item \textbf{Geometric Resolution:} 
    To construct mathematical constraints, we aim to convert geometric relationships between irregular part meshes into constraints defined over surfaces. Specifically, for each part, we first compute its bounding box (bbox) in canonical pose. For each bbox face, we then search on the mesh surface for all connected face patches whose normals are aligned with the normal of that bbox face. Among these candidate patches, we select the one that is closest to the bbox face and has the largest area. We designate it as the directional surface for that face. If no suitable mesh patches can be found, we directly use the bbox face itself as the directional surface. In this way, each part is associated with six directional surfaces corresponding to the six faces of its bounding box.

\end{itemize}

\begin{figure*}[tbp]
    \centering
    \includegraphics[width=1\linewidth]{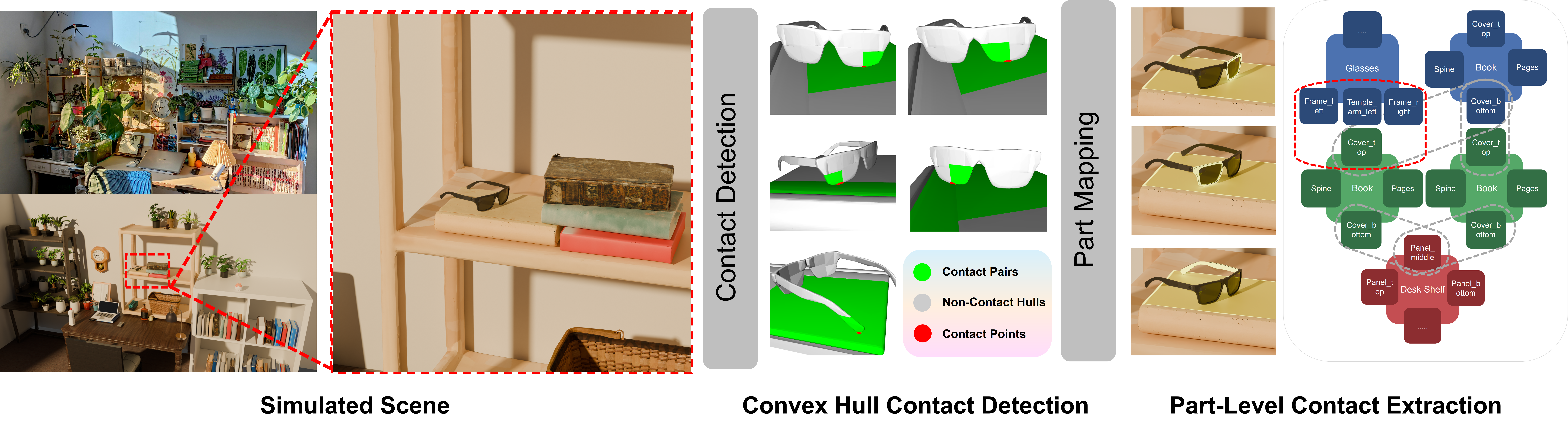}
    \caption{\textbf{Overview of the part-level contact detection pipeline.} After simulating the scene, we perform contact detection on collision meshes by uniformly sampling surface points on every pair of convex hulls across object pairs. Each detected convex-hull contact pair is then mapped back to its corresponding visual-mesh parts, producing a set of part-level contact pairs (each row maps its hull-level pairs to the corresponding final part-level pair on the right). These part contacts are aggregated to construct the part-level contact graph (the red highlight indicating the contacts between the glasses and the book).}
    \label{fig:contactdec}
\end{figure*}

\paragraph{Coarse Localization (Object-Level).}
This stage focuses on object-level relational constraints, which establish coarse spatial arrangements between objects and thereby reduce the search space for final feasible poses. We initiate the process by identifying the parent's support surface $\mathcal{S}_{supp}$ and subtracting the regions already occupied by other objects to obtain the initial collision-free space $\mathcal{S}_{free}$.
Subsequently, we systematically apply the Object-Level Coarse Relations (see \cref{tab:relation_definitions}) to further contract this region. Directional relations (\eg, \texttt{SIDE\_OF}) are implemented by specifying the relative positions of object centroids; proximity relations (\eg, \texttt{NEAR}, \texttt{FAR}) specify inclusion or exclusion regions based on prescribed distances; and alignment relations (\eg, \texttt{ALONG}) constrain objects to lie on the same line; and orientation relations (\eg, \texttt{TOWARD}) impose a coarse ``Look-At'' constraint that binds the object's forward vector to the target's centroid. Based on these 3D constraints, we further filter the original $\mathcal{S}_{free}$ to obtain the final reduced space for this stage, denoted as $\mathcal{S}_{coarse}$.

\paragraph{Fine-Grained Alignment (Part-Level).}
Building upon the resolved surfaces, this phase further determines the object's full 6-DoF pose based on part-level constraints. For every part node on the object that participates in at least one relation edge, we sequentially solve its associated constraint. For each surface of each part, we compute its feasible pose space by solving the relation constraints with the corresponding surface of the other object's part. The intersection of the result pose spaces of all six surfaces is then taken as the valid pose space for that part; if this intersection is empty, the object is immediately marked as a failure case, and we proceed to the next object. After processing all relation-involved parts, we intersect their individual pose spaces with $\mathcal{S}_{coarse}$ to obtain the object’s final feasible pose space for this stage, marked as $\mathcal{S}_{fine}$. If this final intersection is empty, the object is likewise labeled as a failure case and discarded before moving on.

\paragraph{Sampling and Validation.}
To finalize the object's placement, we uniformly sample a candidate pose $\mathbf{t}$ from the minimal subspace $\mathcal{S}_{fine}$. This candidate undergoes a validation procedure to ensure physical and semantic consistency. First, we verify adherence to all active geometric constraints defined in \cref{tab:relation_definitions}, ensuring that the generated pose satisfies the full set of relational requirements. Second, we perform a global collision check against the static scene using FCL-accelerated mesh-level interference detection. If a candidate fails validation, it is discarded, and we resample a new pose; if the process exceeds a predefined maximum number of trials without success, the placement is flagged as a failure.


\section{Details of Asset Library}

Our dataset is constructed by combining a subset of PartNeXt~\cite{wang2025partnext} data, which provides part annotations over integrated assets from 3D-FRONT~\cite{fu20213dfront}, ABO~\cite{collins2022abo}, and Objaverse~\cite{deitke2023objaverse}. In addition, for object categories that commonly appear in daily environments but are not covered by PartNeXt, we first collect reference images from the Internet and use Rodin~\cite{zhang2024clay} to synthesize corresponding 3D assets. We then apply P3-SAM~\cite{ma2025p3} to perform part segmentation and use GPT~\cite{openai2025gpt5} to generate part-level semantic annotations. Finally, we manually verify and adjust the canonical pose of instances in each object category to ensure category-level pose consistency, which is crucial to applying relation constraints in the subsequent solver.

\begin{figure}
    \centering
    \includegraphics[width=1\linewidth]{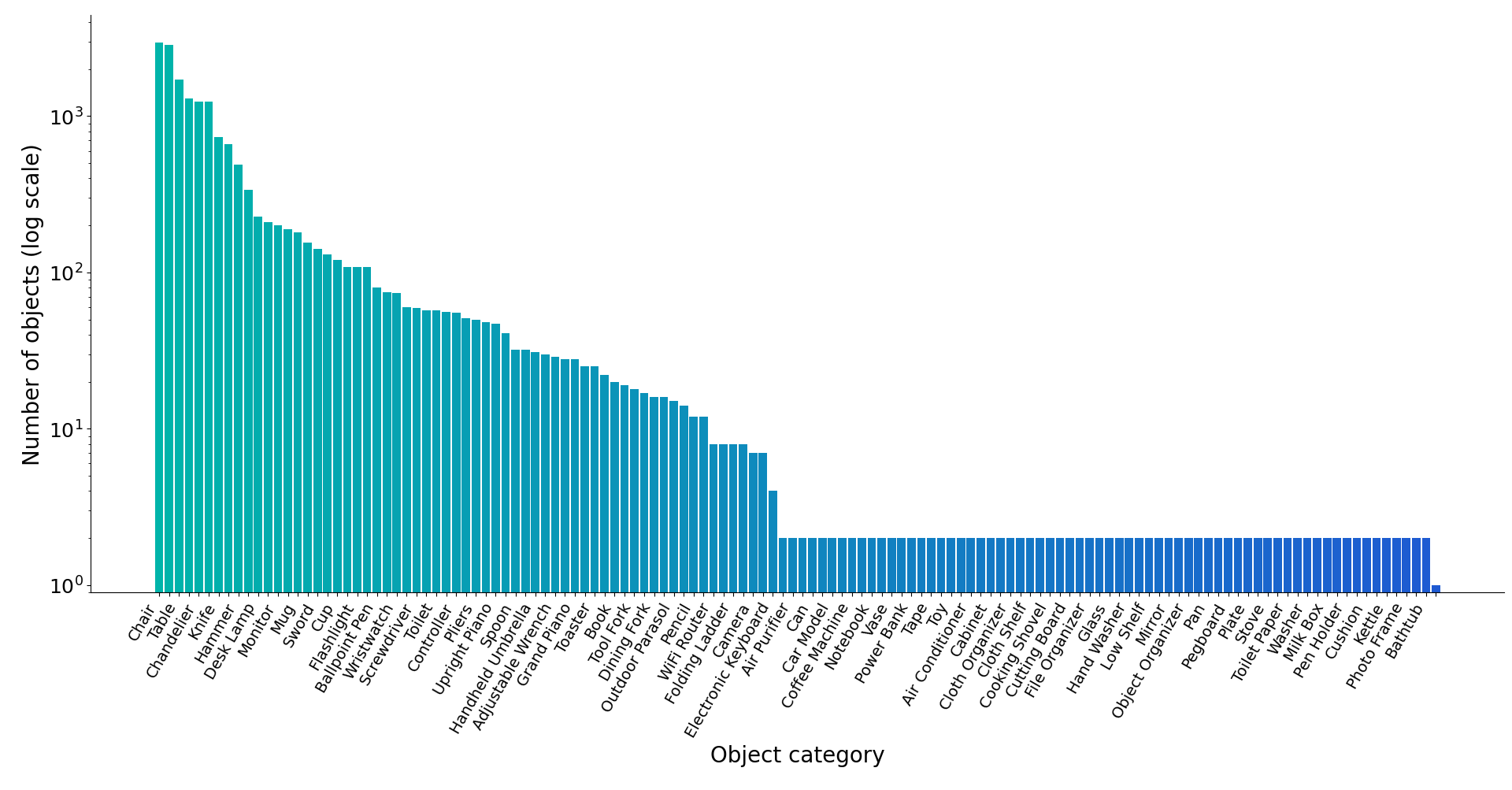}
    \caption{Distribution of Object Categories}
    \label{fig:objcate}

    \centering
    \includegraphics[width=1\linewidth]{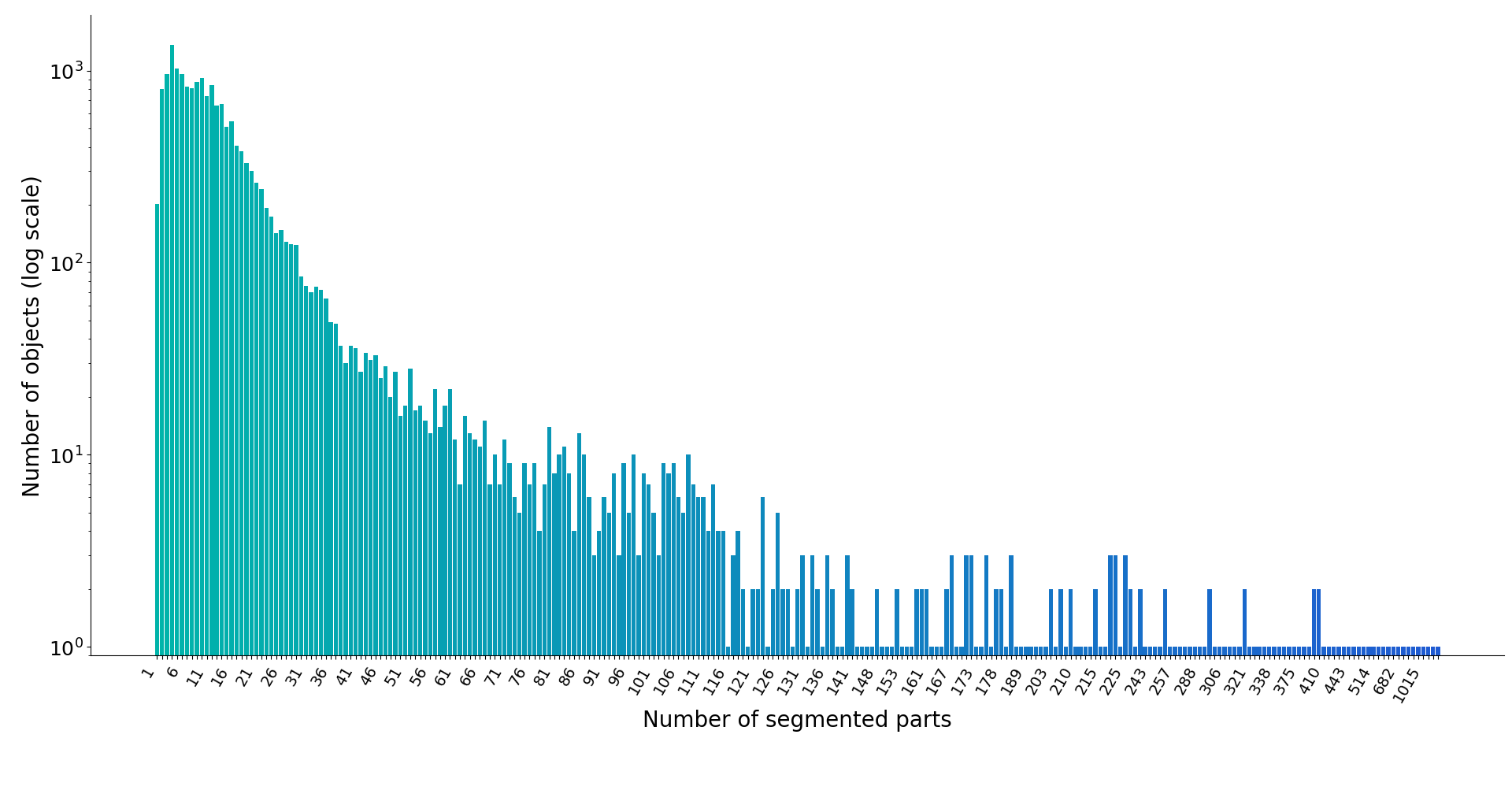}
    \caption{Distribution of Number of Parts Count per Object}
    \label{fig:partdis}
    \centering
    \includegraphics[width=0.9\linewidth]{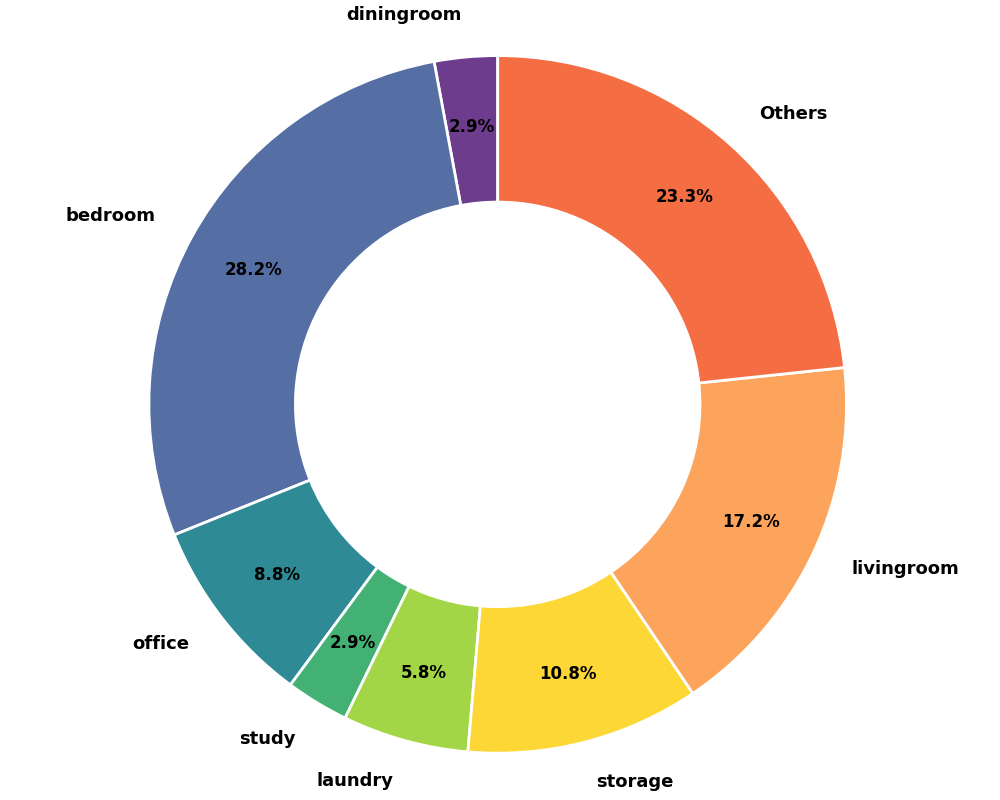}
    \caption{Distribution of Room Types}
    \label{fig:roomtype}
\end{figure}

To better characterize the properties of our assembled asset collection, we perform a detailed statistical analysis of both object categories and the number of part segments per object. As shown in \cref{fig:objcate}, the category composition reveals a broad coverage of everyday object types. In addition, \cref{fig:partdis} presents the distribution of segmented part counts per object. On average, each object contains 17.4 parts.

\section{Scenes}

\subsection{Scene Simulation Procedure and Contact Extraction}
After generating each scene, we simulate it in Sapien~\cite{xiang2020sapien} and subsequently extract the part-level contact relations. Concretely, we first apply COACD~\cite{wei2022approximate} to perform convex decomposition for every object. During simulation, we use the COACD-generated meshes as collision meshes for stable and accurate physical interactions, while the original meshes are kept only for visualization. The simulation runs with a timestep of 0.01~s for 1000 steps, and all objects are assigned a linear damping of 0.05~m/s and an angular damping of 0.5~deg/s. After simulation, we extract the contact graph of the entire scene, and the whole process is shown in \cref{fig:contactdec}. First, we perform contact detection by sampling points on the surfaces of each convex hull of each collision mesh. Next, for each pair of objects, we examine every pair of their convex hulls. For both convex hulls in each pair, we uniformly sample surface points and record all points whose distance to the opposite hull is below 0.001 m. Each recorded point is then mapped to the nearest visual-mesh part on its own object, producing a pair of contacting parts corresponding to the two convex hulls. By aggregating all such part pairs across the scene, we construct an undirected graph that represents the part-level contact graph of the entire environment.

\begin{figure}
    \centering
    \includegraphics[width=\linewidth]{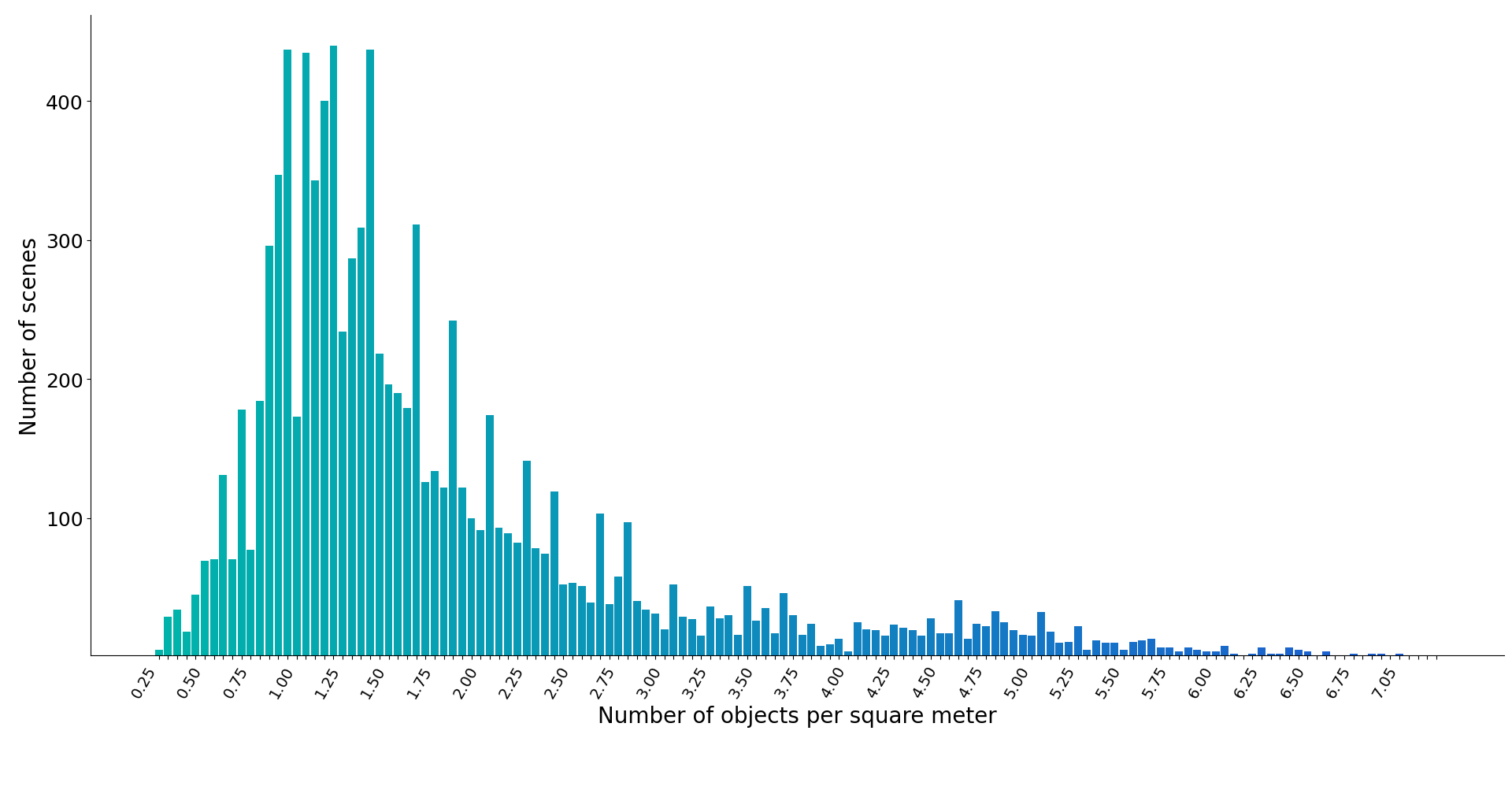}
    \caption{Distribution of Object Density Across Rooms}
    \label{fig:objden}

    \centering
    \includegraphics[width=\linewidth]{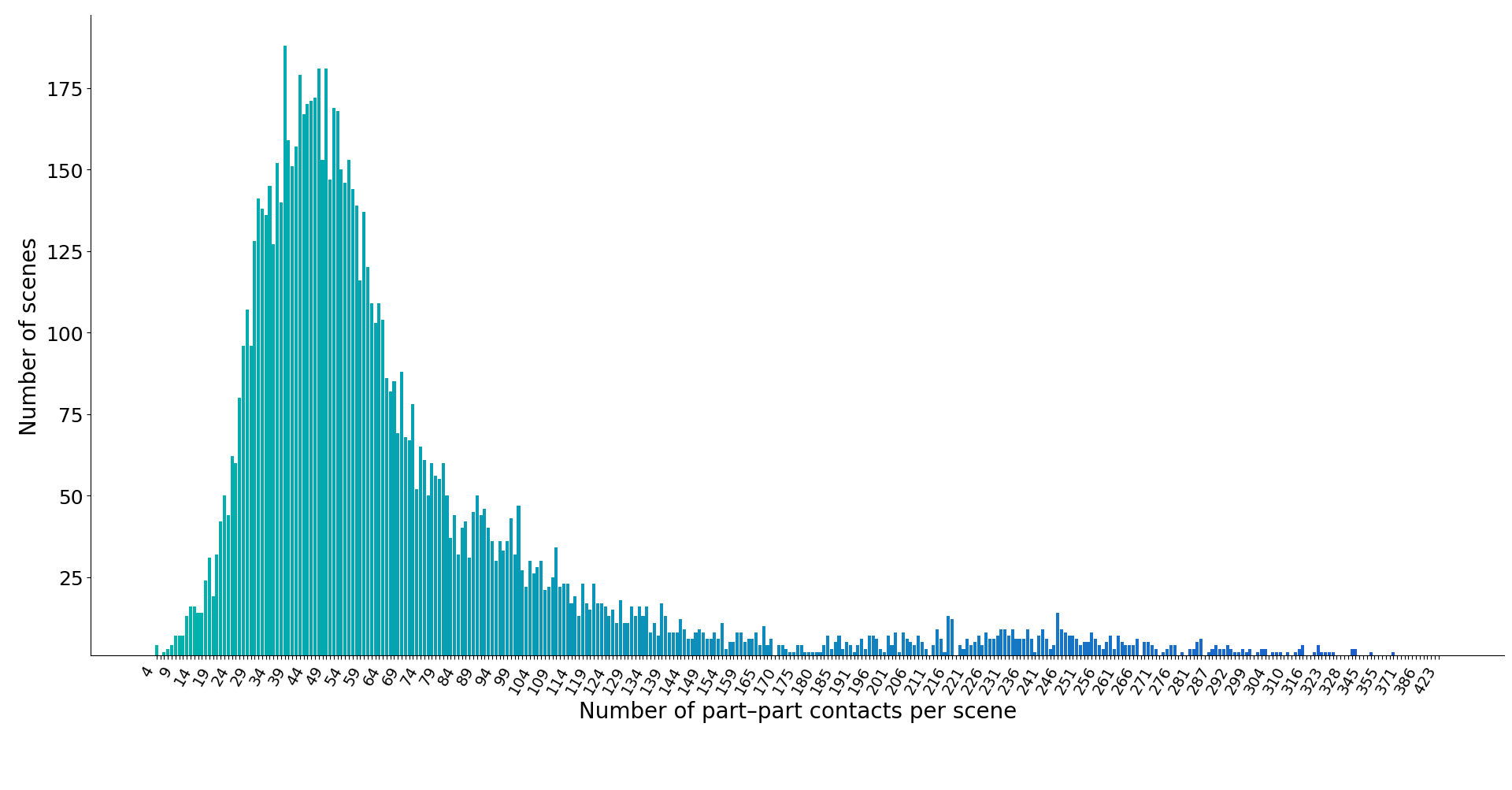}
    \caption{Distribution of Part–Part Contacts per Scene}
    \label{fig:contactsta}
\end{figure}

\subsection{Scene Dataset Statistics and Analysis}
\begin{figure}
    \centering
    \includegraphics[width=\linewidth]{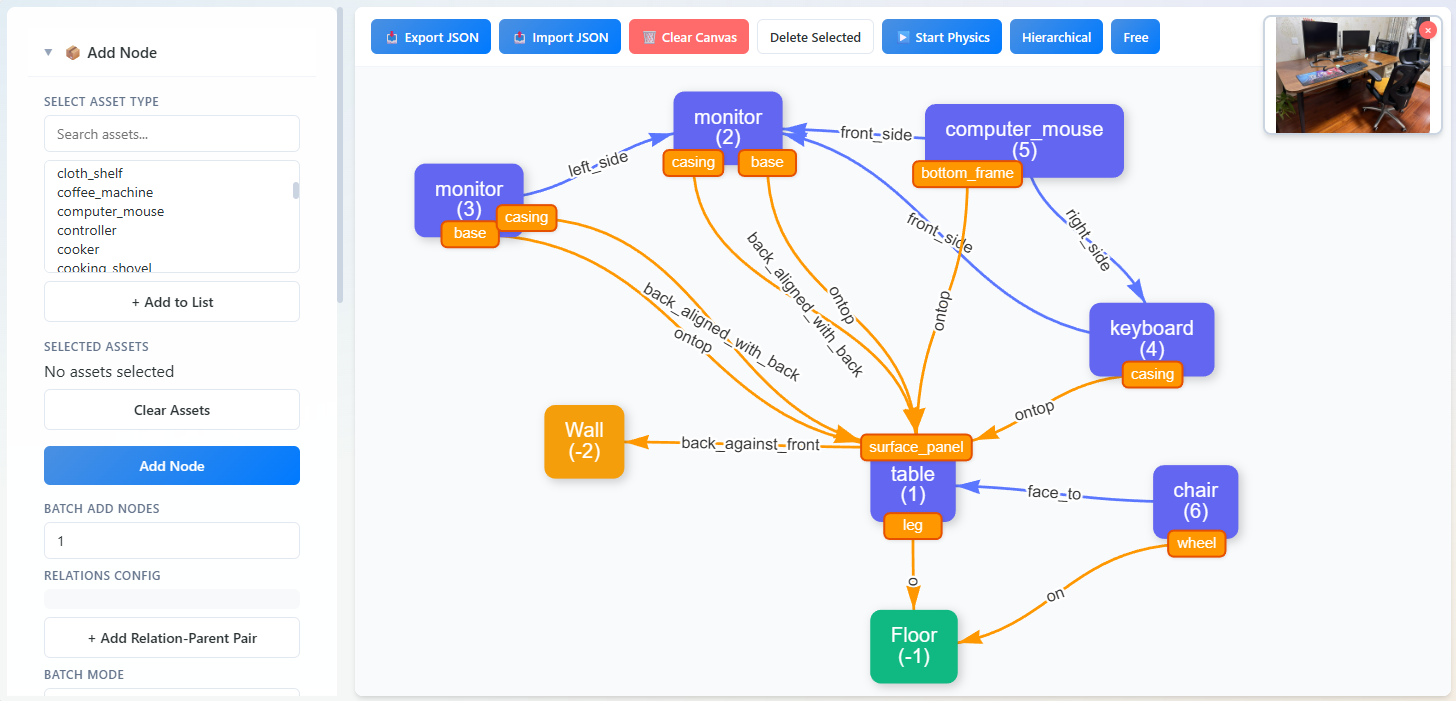}
    \caption{User Interface for Designing PAGs}
    \label{fig:ui}
\end{figure}

Our scene data are constructed as follows. With our UI shown in \cref{fig:ui}, we first design 200 distinct PAGs inspired by real-world images and categorize them into 17 room types, including bedroom, living room, office, study, laundry, storage, dining room, bathroom, tea house, game room, kitchen, music room, bar, dormitory, meeting room, hotel, and grocery. These PAGs collectively drive the generation of 10,000 scenes, providing broad coverage of diverse and commonly encountered indoor environments. The distribution of these scenes across different room types is illustrated in \cref{fig:roomtype}. To validate the quality of our generated scenes, we aim to show that they are both structurally complex and rich in physical interactions. To this end, we compute the object density (number of objects per square meter) for each room, as shown in \cref{fig:objden}. We find that more than 55\% of the scenes have an object density exceeding 1.35 objects per square meter. To demonstrate the richness of physical interactions, we analyze the distribution of the number of contacts per scene of our generated scenes, shown in \cref{fig:contactsta}. The results show that more than 57\% of the scenes contain more than 50 part-level contacts, making them ideal for studying complex, multi-object interaction and physical stability.

\section{Details of Experiments on Spatial Tasks}

\subsection{Implementation Details of VLM Spatial Reasoning}

We fine-tune a large vision language model on 3D scene understanding tasks using the \textbf{Qwen3-VL-235B-A22B-Instruct}~\cite{qwen2025qwen3vl}, which is a 235B-parameter, instruction-tuned multimodal model capable of jointly processing images and text. To make training tractable at this scale, we adopt a parameter-efficient LoRA~\cite{hu2022lora} strategy rather than full-parameter fine-tuning: low-rank adapters are inserted into all linear layers of the transformer blocks, with rank~8 and scaling factor~32, and only these adapter parameters are updated. The underlying pre-trained weights of Qwen3-VL remain frozen throughout fine-tuning.

Images are fed through the official Qwen3-VL visual encoder at a fixed maximum resolution, with the number of visual tokens capped at $1024$ (via \texttt{IMAGE\_MAX\_TOKEN\_NUM}). Pre-processing follows the original resizing and padding pipeline, after which the image tokens are concatenated with text tokens into a single multimodal sequence. For each 2D view rendered from the 3D scenes, we project the 3D object boxes to the image plane and obtain 2D bounding boxes in pixel coordinates. These coordinates are normalized and quantized before being injected into the text. Concretely, a coordinate $(x,y)$ is divided by the image width and height, respectively, mapped into the range $[0,1]$, and then discretized into integer bins in $[0,1000]$. A bbox is serialized as a short textual span \texttt{<|box\_start|>x\_min, y\_min, x\_max, y\_max<|box\_end|>}, which is then tokenized by the standard tokenizer. This encoding makes spatial information explicit while remaining compatible with the instruction-tuned interface of the base model.

The model is trained in a multi-task fashion on three types of supervision: visual relation multiple-choice questions, part-level contact multiple-choice questions, and scene graph generation. In the visual relation MCQ task, each training example selects a pair of objects $A$ and $B$ in the image, provides their bounding
boxes highlighted on the image, and asks for their relationship (for example ``left of'', ``on'' or ``no contact''). The ground-truth option is determined from the underlying 3D annotations, and the assistant response is supervised to be exactly the letter of the correct option.

A typical prompt for the visual relation MCQ task is:
\begin{quote}
\small
\textbf{User:} The image displays two objects highlighted with a red and a blue bounding box. Your task is to identify the object within the red box, the object within the blue box, and the spatial relationship between them. From the options below, choose the one that correctly describes this scene. (A) \dots, (B) \dots, (C) \dots, (D) \dots. Your answer should only be the letter of the correct option.\\
\textbf{Assistant:} C
\end{quote}
The part-level contact MCQ task refines this setup by asking which specific parts of two objects are in physical contact, rather than only their coarse relation. Each example highlights two objects (for instance, a bookcase and a bucket) with red and blue boxes in the image, and the prompt enumerates several hypotheses involving part names drawn from a predefined part ontology (\eg, ``Shelf Body'' or ``Back Panel'' for the bookcase, ``Main Body'' or ``Handle'' for the bucket), together with a ``no contact'' option. The correct option is derived from part-level contact labels computed in the 3D space, and the model is again trained to output only the option letter.

A typical prompt for the part-level contact MCQ task is:
\begin{quote}
\small
\textbf{User:} The image displays two objects highlighted with a red and a blue bounding box. Your task is to identify the object within the red box, the object within the blue box, and whether they are in contact with each other. From the options below, choose the one that correctly describes this scene. (A) \dots, (B) \dots, (C) \dots, (D) \dots. Your answer should only be the letter of the correct option.\\
\textbf{Assistant:} C
\end{quote}
Finally, the scene graph generation task supervises the model to produce a global relational description of the scene. Here the user message lists the object  and relation candidates and asks the model to detect and localize all objects, then identify the spatial relationships between the detected objects. The assistant is trained to return a complex scene graph, such as a sequence of ``(bookcase, supporting, cup); (table, under, lamp); \dots'' triplets. Unlike the MCQ tasks, this supervision is open-ended and encourages the model to maintain precies analysis over the whole scene.

All tasks are packed into a unified JSON-based dialogue format, where each entry contains a \texttt{messages} field (a short user–assistant conversation) and a list of associated images. The task type is encoded purely in natural language, without special control tokens. During fine-tuning we simply mix the three datasets according to fixed sampling ratios and train a single set of LoRA adapters on the combined corpus. Optimization is performed with AdamW at a learning rate of $1\times 10^{-4}$ with linear warm-up over $5\%$ of the updates and a minimum learning rate of $1\times 10^{-5}$, using a global batch size of 16 (micro-batch size 2 per GPU on 8 NVIDIA H20 GPUs). We run one epoch over the mixed dataset, perform evaluation on held-out validation splits corresponding to each task, and report final results on disjoint test sets sampled at the scene level to avoid leakage across different camera views.

\subsection{Implementation Details of Scene Generation}

We represent each scene as an object-centric graph. Each object node carries its pose parameters (translation with 3 DOF, scale with 3 DOF, and rotation with 6 DOF), forming a 12-dimensional geometric descriptor, and a Michelangelo-based shape embedding~\cite{zhao2023michelangelo} (described below). The geometric descriptor and positional code are concatenated and projected into the 512-dimensional model space by a linear layer. In parallel, for each object we extract a 256 $\times$ 64-dimensional latent code from the pretrained Michelangelo encoder, which maps 3D shapes into a joint shape–image–text latent space. This latent is again projected to 512 dimensions through a linear layer and interacts with the pose representation inside the multimodal graph blocks, providing high-level priors about plausible object shapes and categories.

Pairwise relations between objects are modeled as directed edges in the scene graph. For every ordered pair of nodes, we associate a CLIP~\cite{Radford2021LearningTV} relation embedding in $\mathbb{R}^{768}$, which captures semantic relations such as ``left of'', ``below'', or ``supporting''. These 768-dimensional vectors are processed by a two-layer MLP that maps them into the 512-dimensional edge space used by the Graph Transformer~\cite{dwivedi2021generalization}. Inside each block, graph attention operates jointly on nodes and edges: attention logits between nodes are modulated by the corresponding edge embeddings, and node and edge streams are updated by separate feed-forward networks. All submodules are wrapped with residual connections and either standard LayerNorm~\cite{ba2016layernormalization} or AdaLayerNorm~\cite{perez2018film}, depending on whether timestep conditioning is required.

We treat 3D layout prediction as a conditional diffusion process~\cite{10.5555/3495724.3496298} over object poses. The forward noising process follows the DDPM formulation, using 1000 diffusion steps and a squared cosine beta schedule. At each training step, we sample a timestep $t$ uniformly from 1 to 1000, add Gaussian noise to the ground-truth boxes according to the DDPM variance schedule, and train the network to predict the added noise (epsilon prediction). The timestep $t$ is first converted into a sinusoidal embedding and then mapped to the 128-dimensional vector used by AdaLayerNorm; this embedding is broadcast to all layers and injected before graph attention and before the feed-forward networks, so that the model learns a time-dependent denoising trajectory in the space of 3D layouts.

For optimization, we use AdamW, with the learning rate set to $8\times 10^{-4}$, the weight decay to $1\times 10^{-4}$, and the Adam betas to $(0.9, 0.999)$. Training is performed on eight NVIDIA H20 GPUs using the \texttt{accelerate} library, with a per-GPU batch size of a single scene graph. For each scene in the batch, we further sample 16 independent noise vectors and timesteps, resulting in 16 noisy versions of the same ground-truth layout; this yields an effective batch size of 128 diffusion samples per optimization step across all GPUs. We train for 2000 epochs, with 500 gradient updates per epoch. Throughout training, we maintain an exponential moving average of all model parameters with a maximum decay rate of 0.9999 and a warm-up strategy controlled by inv\_gamma $= 1.0$ and power $= 0.75$, following the default configuration used in \texttt{diffusers}. All quantitative metrics and qualitative visualizations reported in the paper are obtained using these EMA-smoothed weights at test time.



\end{document}